\documentclass[letterpaper]{article} 
\usepackage{amssymb}
\usepackage{aaai2026}  
\usepackage{times}  
\usepackage{helvet}  
\usepackage{courier}  
\usepackage[hyphens]{url}  
\usepackage{graphicx} 
\usepackage{natbib}  
\usepackage{caption} 
\usepackage{booktabs, siunitx, adjustbox}

\usepackage{amsmath} 
\usepackage{algorithm}
\usepackage{algorithmic}
\usepackage{enumitem}
\usepackage{newfloat}
\usepackage{listings}
\DeclareCaptionStyle{ruled}{labelfont=normalfont,labelsep=colon,strut=off} 
\floatstyle{ruled}
\newfloat{listing}{tb}{lst}{}
\floatname{listing}{Listing}
\usepackage[utf8]{inputenc}
\usepackage{newunicodechar}
\newunicodechar{【}{[}
\newunicodechar{】}{]}

\usepackage{booktabs}      
\usepackage{multirow}      
\usepackage{siunitx}       
\usepackage{pifont}        
\usepackage{caption}

\usepackage[utf8]{inputenc} 
\usepackage{booktabs}   
\usepackage{tabularx}   
\usepackage{array}      
\usepackage{pifont}     

\title{KnowDR-REC: A Benchmark for Referring Expression Comprehension with Real-World Knowledge}

\author{
  Guanghao Jin\textsuperscript{*}\textsuperscript{\rm 1,2},
  Jingpei Wu\textsuperscript{*}\textsuperscript{\rm 2},
  Tianpei Guo\textsuperscript{\rm 1},
  Yiyi Niu\textsuperscript{\rm 3},
  Weidong Zhou\textsuperscript{\rm 1},
  Guoyang Liu\textsuperscript{\dag}\textsuperscript{\rm 1}
}
\affiliations{
  \textsuperscript{\rm 1}Shandong University \quad
  \textsuperscript{\rm 2}LMU Munich \quad
  \textsuperscript{\rm 3}Technical University of Munich\\
  \ G.Jin@campus.lmu.de, jingpei.wu@lmu.de, 202200201042@mail.sdu.edu.cn,\\
  yiyi.niu@tum.de, \{wdzhou, gyliu\}@sdu.edu.cn
}

\begin{document}
\nocopyright            

\maketitle

\renewcommand{\thefootnote}{\fnsymbol{footnote}}
\footnotetext[1]{Equal contribution.}
\footnotetext[2]{Corresponding author.}
\renewcommand{\thefootnote}{\arabic{footnote}}

\begin{figure*}[t]
  \centering
  \includegraphics[width=1\textwidth]{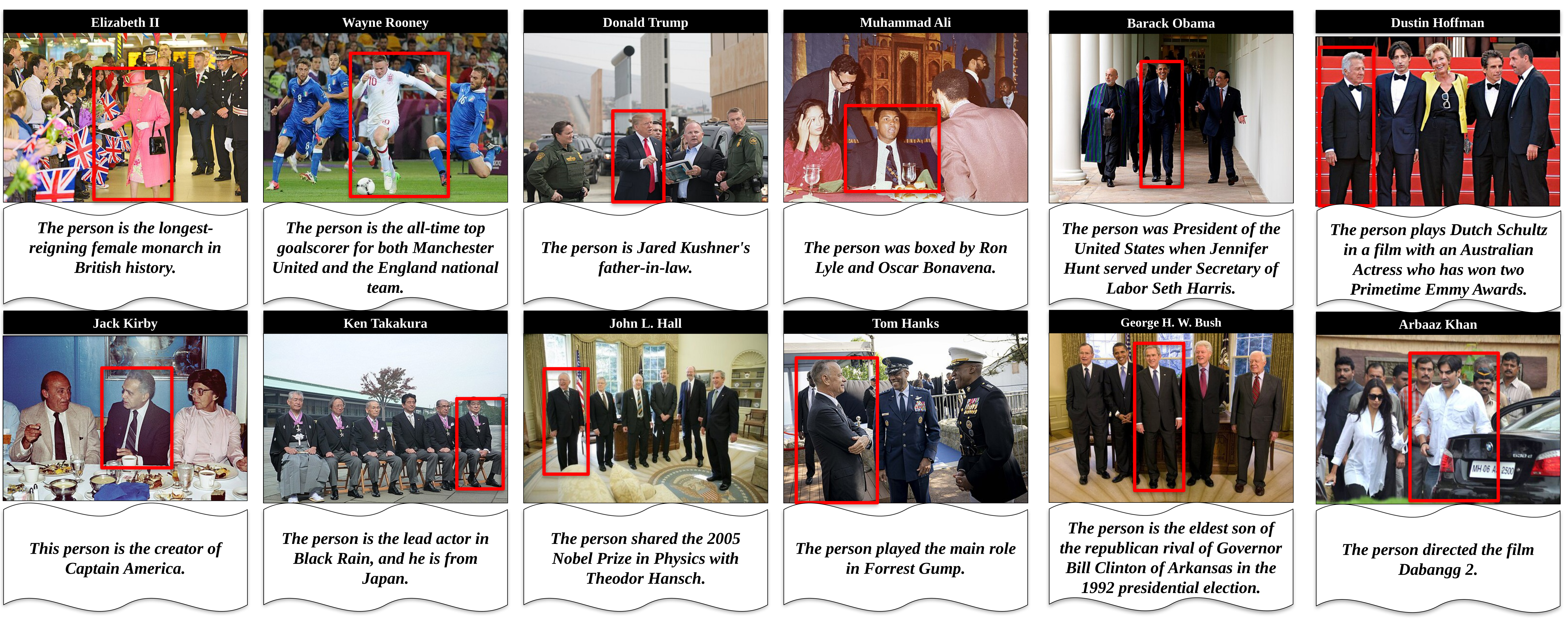}
  \caption{KnowDR-REC is a novel benchmark that aims to integrate external world knowledge into the textual expressions used in referring expression comprehension (REC).}
  \label{fig:1}
\end{figure*}

\begin{abstract}
Referring Expression Comprehension (REC) is a popular multimodal task that aims to accurately detect target objects within a single image based on a given textual expression. However, due to the limitations of earlier models, traditional REC benchmarks either rely solely on intra-image cues or lack sufficiently fine-grained instance annotations, making them inadequate for evaluating the reasoning capabilities of Multi-modal Large Language Models (MLLMs). To address this gap, we propose a new benchmark, KnowDR-REC, characterized by three key features: Firstly, it is built upon real-world knowledge, requiring fine-grained multimodal reasoning across text and image. Secondly, the dataset includes elaborately constructed negative samples via fine-grained expression editing, designed to evaluate a model's robustness and anti-hallucination ability. Lastly, we introduce three novel evaluation metrics to systematically explore the model's internal reasoning process. We evaluate 16 state-of-the-art multimodal models on KnowDR-REC, with experimental results showing that existing MLLMs still struggle with knowledge-driven visual grounding tasks. Furthermore, we observe a decoupling between textual understanding and visual grounding in MLLMs, where many models are significantly influenced by memorized shortcut correlations, which severely affect their behavior on our benchmark and hinder genuine multimodal reasoning. We anticipate that the proposed benchmark will inspire future research towards developing more robust, interpretable, and knowledge-intensive visual grounding frameworks, driving the development of more reliable and robust multimodal systems for complex real-world scenarios.


\end{abstract}

\begin{links}
   \link{Code}{https://github.com/LetItBe12345/KnowDR-REC}

\end{links}

\section{Introduction}

The advent of Multi-modal Large Language Models (MLLMs) has significantly transformed artificial intelligence research, particularly impacting tasks that require joint vision-language understanding \cite{liu2023visual, huang2023language, zang2025contextual}. Among these tasks, Referring Expression Comprehension (REC) \cite{qiao2020referring} stands out as a core challenge, aiming to detect target objects in images based on natural language descriptions. Despite considerable progress, existing REC evaluation paradigms remain insufficient for adequately reflecting the inherent complexity of real-world applications. To bridge this gap, we introduce Knowledge-Driven Referring Expression Comprehension (KnowDR-REC), a novel benchmark explicitly designed to evaluate MLLMs on knowledge-intensive visual referring tasks. Representative examples from this benchmark are illustrated in Figure~\ref{fig:1}. 

Traditional REC benchmarks predominantly focus on visual cues, spatial relations, and simple attribute matching within images \cite{liu2019clevr}. Widely-used benchmarks like RefCOCO \cite{yu2016modeling}, RefCOCO+ \cite{yu2016modeling}, and RefCOCOg \cite{mao2016generation} have established foundational standards but significantly underestimate the complexity of natural human referential behaviors. Human references typically integrate external knowledge across temporal, cultural, and factual dimensions.  For instance, a seemingly straightforward phrase like “the flag bearer wearing the Chinese delegation uniform from the 2024 Paris Olympics” actually requires deep chains of external knowledge, including temporal references (2024), events (Paris Olympics), national affiliation (Chinese delegation), and roles (flag bearer). Current REC benchmarks fail to cover such reasoning-intensive tasks, leaving a substantial gap in evaluating the ability of MLLMs to integrate external knowledge and perform multi-hop reasoning.  

Another major limitation of existing Referring Expression Comprehension (REC) benchmarks is the absence of fine-grained target annotations \cite{wu2020phrasecut}. Most current datasets provide only coarse-grained descriptions such as “the woman in red” or “the man wearing a hat,” which are insufficient for evaluating fine-grained perception in real-world scenarios. In practical scenarios, models must handle highly specific referring expressions, such as “the 2022 Nobel laureate in Physics”. These tasks not only challenge models' fine-grained visual grounding capabilities but also test their capacity for cross-modal reasoning. Evidently, existing benchmarks fall short of adequately evaluating MLLM performance under such fine-grained conditions.

Moreover, current REC benchmarks largely neglect unanswerable cases. While recent frameworks have introduced negative samples to assess refusal capabilities of models \cite{liu2024finecops,schulter2023omnilabel}, they often lack in-depth analysis of model errors. As a result, it remains difficult to evaluate a model’s robustness and anti-hallucination ability in ambiguous or invalid scenarios.

The proposed KnowDR-REC benchmark advances REC evaluation along three key dimensions. First, we introduce external world knowledge, requiring models to perform multi-hop reasoning at both the textual and visual levels. In addition, we propose novel metrics designed to probe the models' internal reasoning processes. Second, our benchmark emphasize fine-grained instance grounding, particularly targeting human subjects annotated with detailed attributes and contextual information. Third, we leverage temporal knowledge graphs \cite{cai2022temporal} to perform fine-grained editing of referring expressions, enabling diverse negative samples to assess model robustness and hallucination resistance.

To validate the benchmark's effectiveness, we conduct a comprehensive evaluation across 16 models, covering three categories: (1) closed-source generalist MLLMs; (2) open-source generalist MLLMs; and (3) open-source specialist MLLMs designed for visual grounding. Both conventional metrics and new metrics tailored for knowledge-intensive settings are employed. Experimental results reveal several critical issues with current multimodal models. Firstly, the models exhibit unsatisfactory performance on both positive and negative samples, indicating limitations in their textual grounding, visual grounding and anti-hallucination capabilities. Secondly, although textual reasoning accuracy is highly correlated with visual grounding correctness, models surprisingly demonstrate a certain level of grounding ability even when their textual reasoning is flawed. Most alarmingly, models often exhibit similar grounding behaviors regardless of the correctness of textual expressions, suggesting they rely not solely on the provided referring expressions but also heavily on memorized shortcuts. This phenomenon raises significant concerns about the reliability and robustness of these models in knowledge-intensive visual grounding tasks.

Our main contributions are as follows:

\begin{enumerate}[label=(\arabic*)]

    \item We propose KnowDR-REC, the first benchmark explicitly designed to assess MLLMs on knowledge-driven, multi-hop reasoning in referring expression comprehension.
    \item We develop a temporal knowledge graph-based negative sampling strategy to rigorously evaluate the robustness and anti-hallucination capabilities of MLLMs.
    \item We introduce novel conditional evaluation metrics to precisely measure multimodal models' textual reasoning and visual grounding performance.
    \item Through extensive evaluations of 16 state-of-the-art MLLMs, we identify critical shortcomings of LLMs including reliance on memorized shortcuts and weak robustness against semantically subtle negative samples.
    
\end{enumerate}

\section{Related Work}

\subsection{Referring Expression Comprehension}

Referring Expression Comprehension (REC) is a core task in multimodal understanding that detects the visual target based on natural language. Classic benchmarks such as the RefCOCO series rely on short spatial expressions and have been nearly saturated—models like CogVLM \cite{wang2024cogvlm} achieve strong results even in zero-shot settings, with 92.44\% on RefCOCO, 88.55\% on RefCOCO+, and 90.67\% on RefCOCOg (Acc@0.5), revealing a performance ceiling and limited room for further progress.

To advance the task, recent datasets increase complexity: HC-RefLoCo \cite{wei2024large} uses long human-centric expressions, FineCops-Ref \cite{liu2024finecops} focuses on compositional reasoning, ReSeDis \cite{huang2025resedis} introduces cross-dataset and open-world settings, and MC-Bench \cite{xu2024mc} extends REC to the multi-image domain.

Despite progress, existing benchmarks lack hard negatives, temporal cues, and knowledge-based references—e.g., referring to people via historical or biographical facts—limiting real-world applicability. This calls for more realistic, knowledge-intensive REC datasets and evaluation protocols.

\subsection{Knowledge-intensive Understanding}

Integrating external knowledge is essential for reasoning in multimodal tasks, as many real-world questions require factual, temporal, or relational information beyond what is visible in an image. In Natural Language Processing (NLP), benchmarks like HotpotQA \cite{yang2018hotpotqa} and MetaQA \cite{puerto2021metaqa} test multi-hop reasoning across documents or knowledge graphs. In vision, datasets like OVEN \cite{hu2023open}, WIKIPerson\cite{sun2022visual}, and KVQA \cite{shah2019kvqa} require linking visual entities to external knowledge bases, enabling more accurate and context-aware understanding.

In contrast, existing Referring Expression Comprehension (REC) datasets rely mostly on commonsense or intra-image cues. KB-Ref \cite{wang2020give} incorporates basic commonsense knowledge, but lacks integration of rich factual or temporal knowledge. This exposes a key gap in REC evaluation—namely, the inability to test knowledge-intensive reasoning in multimodal grounding tasks.




\subsection{Multimodal Model Evaluation and Hallucination}
Hallucination has become central to evaluating robustness in MLLMs, particularly vision-language models (VLMs) \cite{liu2024survey,bai2024hallucination}. A special case is "unsolvable problem detection (UPD)," where models should ideally abstain due to the absence or mismatch of visual content. Prior studies systematically explored single-object hallucination \cite{li2023evaluating}, multi-object hallucination \cite{chen2024multi}, abstention in multimodal QA scenarios \cite{miyai2024unsolvable} . However, current REC benchmarks, despite constructing negative samples \cite{liu2024finecops}, fail to provide an explanation for the causes of hallucination. To bridge this gap, our benchmark introduces controlled negative samples through temporal knowledge graph manipulations and employs two evaluation settings, enabling a systematic analysis of hallucination triggers within REC tasks.

\begin{figure*}[t]
  \centering
  \includegraphics[width=0.95\textwidth, height=0.4\textheight]{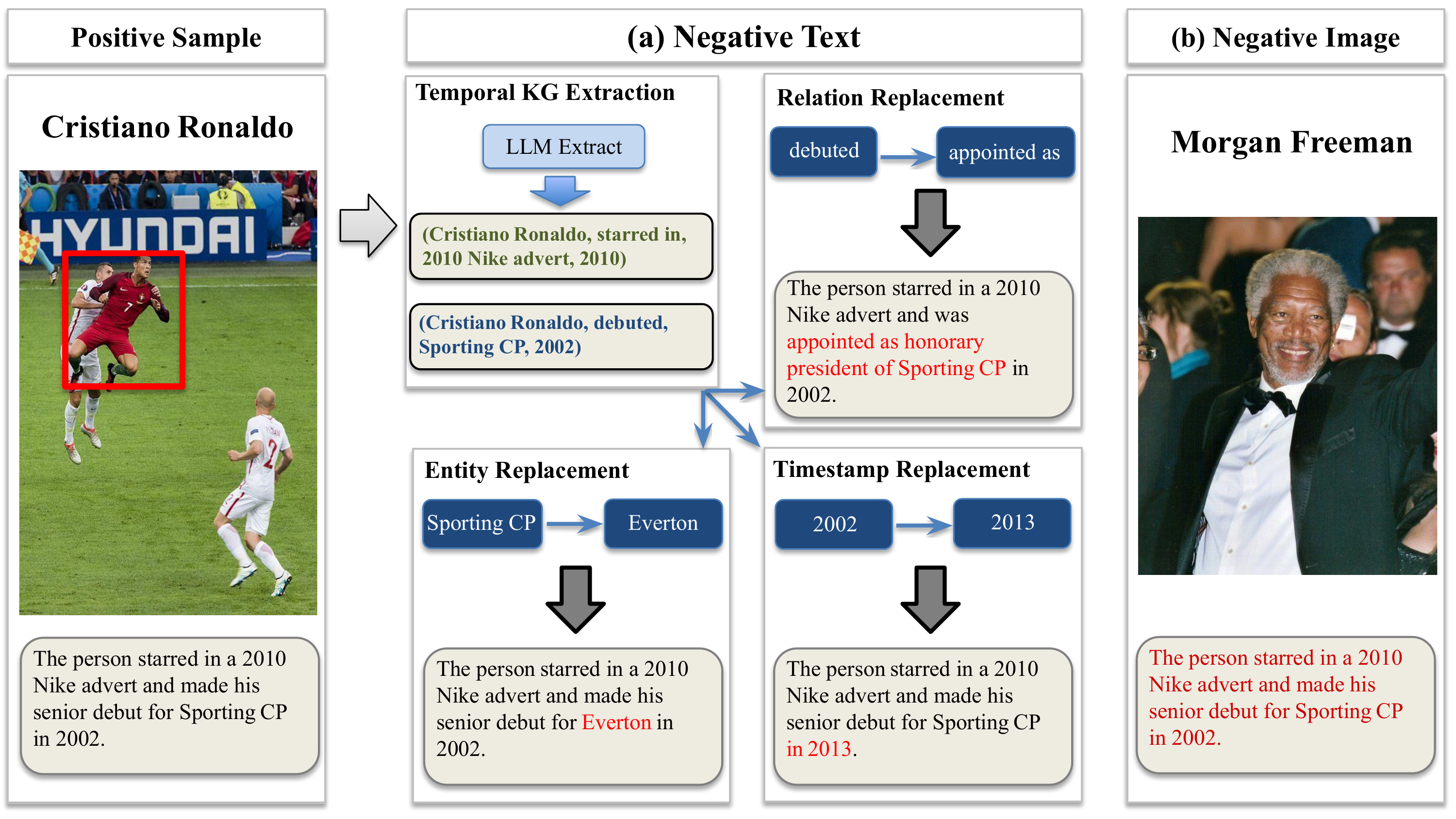}
  \caption{Negative sample construction in KnowDR-REC. Starting from an image-text pair, we extract a temporal knowledge graph from the textual expression. Fine-grained negative samples are introduced by perturbing factual triples within the graph. Additionally, coarse-grained negatives are obtained by pairing the original text with semantically unrelated images using image-text re-ranking.}
  \label{fig:2}
\end{figure*}

\begin{figure}[t]
  \centering
  \includegraphics[width=0.45\textwidth]{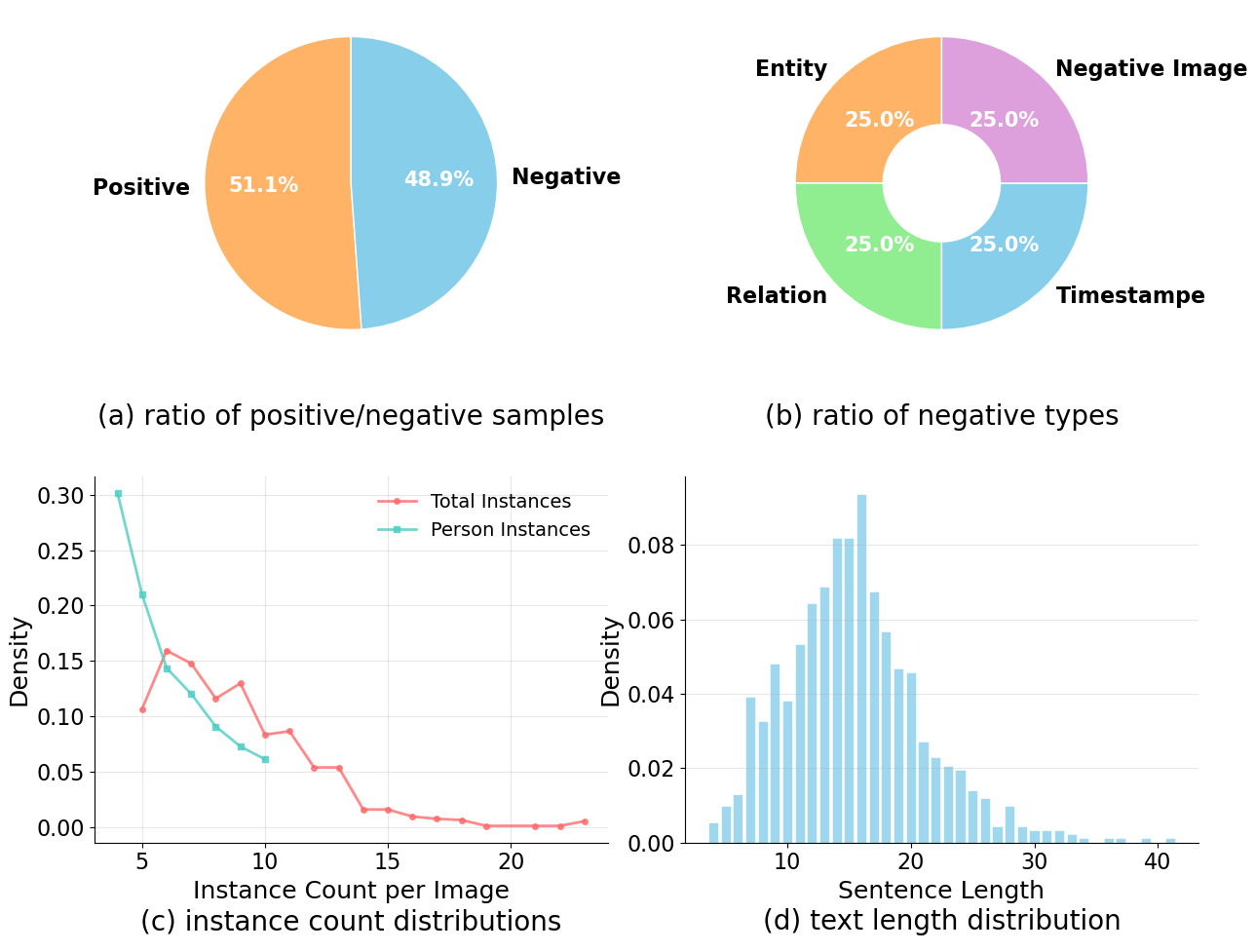}
  \caption{Statistical analysis of the proposed KnowDR-Bench.}
  \label{fig:3}
\end{figure}

\section{Dataset}
In this section, we present the data acquisition process, the annotation pipeline, and the construction procedure for negative samples. The statistics of our dataset are presented in Figure~\ref{fig:3}.

\subsection{Referring Expression Collection}
In contrast to prior REC benchmarks that primarily concern object-level attributes or intra-image spatial relations, KnowDR-Rec focuses on referring expressions grounded in external world knowledge, making their manual construction notably more challenging. Existing knowledge-driven datasets, such as OK-VQA \cite{marino2019ok}, typically depend on a single source and rely on rigid, template-based generation, leading to repetitive phrasing and limited relational diversity—ultimately falling short in supporting rich and diverse reasoning.

To obtain more natural and diverse linguistic expressions, we curate questions from four widely adopted QA datasets in the NLP community---ComplexWebQuestions \cite{talmor2018web}, HotpotQA \cite{yang2018hotpotqa}, KQA Pro \cite{cao2020kqa}, and MetaQA \cite{puerto2021metaqa}---each representing distinct task types. These questions encompass various reasoning paradigms, such as multi-hop reasoning and compositional queries, reflecting a broad spectrum of knowledge and reasoning skills. Notably, since persons are among the most common and richly attributed entity types grounded in external world knowledge, we retain only those questions involving person entities. This facilitates subsequent instance-level annotation using images.

To ensure that each selected question refers unambiguously and precisely to a specific person, we manually filter the candidate set. This step removes samples containing ambiguity, non-unique answers, or vague entity references, ensuring that the final expressions are semantically clear and referentially grounded.

To further improve the naturalness and linguistic variety of the referring expressions, we employ ChatGPT-4o\cite{hurst2024gpt} to transform the original questions into declarative statements tailored to the Referring Expression Comprehension (REC) task setting.

After applying the aforementioned filtering and processing pipeline, we obtain a total of 2,537 referring expressions, which constitute the expression corpus for the KnowDR-REC benchmark.

\subsection{Creating Positive Data}
To support instance-level visual grounding in the REC task, we construct a set of positive samples. We begin by retrieving candidate images from Wikipedia \cite{bridge2001wikipedia} based on the names of the persons referred to in the expressions. To ensure that the images used for grounding and reasoning are relevant and of high quality, we apply a series of filtering criteria. Each image must contain at least four distinct individuals and include instances from more than two different categories. In addition, we require that the image resolution exceeds 300 × 300 pixels to ensure sufficient visual clarity for accurate annotation. All candidate images are processed using YOLOv8 \cite{sohan2024review} to verify compliance with these criteria, ensuring that only visually rich and high-quality images are retained.

We employ two independent annotation teams to perform manual labeling. The first team labels the target by consulting publicly available descriptions of their physical appearance, such as facial features, hairstyle, or clothing, to localize the correct individual in the image. The second team inputs the person’s name and the image into ChatGPT-4o, then uses the model-generated descriptive language to complete the annotation. We compare the results from both annotation strategies and retain only those samples where both teams consistently identify the same visual instance.

As a result, we obtain 1,042 triplets, each consisting of (i) an image, (ii) a declarative referring expression that uniquely identifies a target instance, and (iii) its corresponding ground truth bounding box.

\subsection{Creating Negative Data}
As illustrated in Figure~\ref{fig:2}, to rigorously evaluate the robustness and hallucination resistance of MLLMs on the KnowDR-REC benchmark, we deliberately construct negative samples by introducing perturbations from both textual and visual perspectives to induce semantic deviations.

\paragraph{Negative Expression.}
Inspired by prior approaches for constructing negative samples in referring expression comprehension (REC) benchmark \cite{liu2024finecops}, we generate adversarial expressions via minimal edits that preserve surface-level fluency while introducing referential inconsistencies. These expressions are nearly indistinguishable from positive ones at the surface level, yet semantically incompatible with the visual content. Through such negative samples, we can evaluate the model's fine-grained understanding of expressions as well as its alignment ability with images.

We incorporate the concept of temporal knowledge graphs (TKGs) \cite{cai2022temporal} to support fine-grained semantic modeling and rewriting of natural language expressions. Each expression is parsed into one or more structured tuples of the form:

\[
(s, r, o, t) \quad \text{where} \quad s, o \in \mathcal{E},\ r \in \mathcal{R},\ t \in \mathcal{T}
\]

\noindent Here, $s$ denotes the subject entity, $r$ the relation type, $o$ the object entity, and $t$ the temporal information; $\mathcal{E}$ represents the entity set, $\mathcal{R}$ the relation vocabulary, and $\mathcal{T}$ the set of timestamps. We use GPT-4o to extract these $(s, r, o, t)$ tuples from each declarative sentence, capturing its underlying factual structure. Based on this representation, we construct negative expressions by perturbing a single element of the original tuple. The possible corrupted forms include $(s', r, o, t)$ or $(s, r', o, t)$ or $(s, r, o', t)$ or $(s, r, o, t')$. We then use GPT-4o to rewrite the original sentence according to the perturbed tuple, producing an expression that remains grammatically fluent but is no longer semantically aligned with the visual content. Each generated negative expression is manually verified to satisfy three criteria: (i) minimal surface-level divergence from its corresponding positive counterpart; (ii) factual inconsistency caused by perturbing exactly one element of the tuple; and (iii) a lack of any valid referent within the associated image.

\paragraph{Negative Image.}

In addition to the aforementioned fine-grained negative expression construction, we also adopt a coarse-grained method for generating negative samples—namely, negative image-expression pairs—to evaluate the model's ability to refuse answering when the image and the description are entirely mismatched. Specifically, we randomly shuffle the original image-description pairs and manually verify that the textual description is unrelated to any object present in the randomly selected image.

\section{Experiments}
\subsection{Experimental Setups}
We categorize the evaluated models into three groups: (1) closed-source specialist MLLMs; (2) open-source generalist MLLMs; and (3) open-source specialist MLLMs designed for visual grounding. In this study, we evaluate 16 representative models spanning all these categories. All experiments are conducted under a zero-shot evaluation protocol and run on a computing environment equipped with two NVIDIA RTX 4090 GPUs. The performance results of these models are summarized in Table~\ref{tab:1}.
\subsection{Evaluation Setups}
In knowledge-driven settings, the referring expression comprehension (REC) task relies on two core capabilities: (1) reasoning about the referent based on the input textual expression, and (2) performing visual grounding based on the inferred target. For positive samples, we design a Chain-of-Thought (CoT) prompting strategy\cite{wei2022chain} to explicitly elicit the model's reasoning path. We extract the textual reasoning result from the model’s CoT output. To ensure accurate text-level matching, we construct a character-specific lexicon. Text-level reasoning capability is evaluated using accuracy ($Acc_{Text}$). To assess visual grounding performance, we use the standard accuracy at multiple IoU thresholds (0.50, 0.75, and 0.90), evaluating whether the model’s final answer aligns with the target region. To analyze the impact of textual reasoning on visual grounding, we define two conditional accuracy metrics. Let $B$ denote the event that the predicted bounding box achieves an IoU $\geq 0.5$ with the ground truth, and $N$ denote the event that the model correctly resolves the referential expression at the text level. The conditional metrics are defined as: 
\[
A_{50}^{\text{match}} = P(B \mid N), \quad A_{50}^{\text{mismatch}} = P(B \mid \bar{N})
\]

To evaluate the model’s ability to handle negative samples, we introduce two evaluation protocols: the Basic Setting, which reflects standard REC usage, and the Instruction-guided Setting, which incorporates prompt-based decision-making.

\begin{enumerate}
    \item Basic Setting. In this setting, the model is asked to predict bounding boxes based on the given expression. This corresponds to the conventional Referring Expression Comprehension (REC) task. We use the error rate as the evaluation metric, defined as:

\[
\text{Error Rate} = \displaystyle
\frac{\#~\text{samples with predicted box}}{\#~\text{negative samples}}
\]
    \item Instruction-guided Setting. Here, the model is explicitly instructed to make a binary decision based on the following prompt:  
"Determine whether the instance described in the text exists in the image. If it exists, answer 'yes'; otherwise, answer 'no'."
\end{enumerate}

To avoid inflated performance due to trivial predictions (e.g., always answering 'no'), we include an equal number of positive and negative samples in this setting. The evaluation metric used is accuracy.

\begin{table*}[t]
  \centering
  \footnotesize
  \setlength{\tabcolsep}{5pt}
  \renewcommand{\arraystretch}{1.15}

  \begin{adjustbox}{max width=\linewidth}
  \begin{tabular}{@{}l c *{6}{S} *{2}{S}@{}}
    \toprule
    \multirow{2}{*}{\textbf{Methods}} & \multirow{2}{*}{\textbf{Size}} &
    \multicolumn{6}{c}{\textbf{Positive Sample}} &
    \multicolumn{2}{c}{\textbf{Negative Sample}} \\
    \cmidrule(lr){3-8} \cmidrule(l){9-10}
    & &
    {$\mathrm{Acc}_{0.5}$} &
    {$\mathrm{Acc}_{0.75}$} &
    {$\mathrm{Acc}_{0.9}$} &
    {$\mathrm{Acc}_{Text}$} &
    {$A_{50}^{\text{match}}$} &
    {$A_{50}^{\text{mismatch}}$} &
    {Error Rate} &
    {Acc} \\
    \midrule

    \multicolumn{10}{@{}l}{\textit{Closed-Source Generalist MLLMs}} \\[1pt]
    Gemini 2.5 Flash\cite{comanici2025gemini} & --  & 44.1 & 16.3 & 2.7  & 84.7 & 46.2 & 32.2 & 79.3 & 82.0 \\
    Gemini 2.5 Pro\cite{comanici2025gemini}   & --  & 21.9 & 4.6  & 0.8  & 89.1 & 23.5 & 9.4  & 97.1 & 79.4 \\
    Grok‑2 Vision\cite{xie2025maverix}    & --  & 56.7 & 16.6 & 8.4  & 71.0 & 61.4 & 45.0 & 83.5 & 78.5 \\
    Qwen‑VL‑Plus\cite{bai2023qwenvlversatilevisionlanguagemodel}     & --  & 64.6 & 54.9 & 24.6 & 39.8 & 83.7 & 51.9 & 52.5 & 51.7 \\
    Qwen‑VL‑Max\cite{bai2023qwenvlversatilevisionlanguagemodel}     & --  & 72.2 & 65.6 & 47.2 & 61.4 & 80.4 & 59.0 & 70.7 & 57.4 \\

    \midrule
    \multicolumn{10}{@{}l}{\textit{Open‑Source Generalist MLLMs}} \\[1pt]
    Qwen2.5‑VL\cite{bai2025qwen25vltechnicalreport}       & 3B  & 60.6 & 54.5 & 38.8 & 34.6 & 86.4 & 46.9 & 100.0 & 52.3 \\
    Qwen2.5‑VL\cite{bai2025qwen25vltechnicalreport}       & 7B  & 59.1 & 48.4 & 20.4 & 40.8 & 82.0 & 43.3 & 98.7  & 51.1 \\
    Qwen2.5‑VL\cite{bai2025qwen25vltechnicalreport}       & 32B & 57.6 & 45.9 & 28.3 & 43.2 & 71.6 & 43.8 & 81.3  & 55.5 \\
    Qwen2.5‑VL\cite{bai2025qwen25vltechnicalreport}       & 72B & 71.0 & 63.8 & 45.9 & 61.6 & 68.0 & 56.6 & 81.2  & 54.8 \\
    SPHINX‑Tiny\cite{lin2023sphinx}      & 1.1B  & 8.3  & 2.1  & 1.6  & 10.4 & 21.7 & 6.7  & 100  & 55.1 \\
    SPHINX‑v2‑1k\cite{liu2024sphinx}     & 13B & 38.1 & 14.3 & 3.6  & 32.5 & 42.8 & 35.7 & 93.1  & 61.0 \\

    \midrule
    \multicolumn{10}{@{}l}{\textit{Open‑Source Specialist MLLMs}} \\[1pt]
    VLM‑R1\cite{shen2025vlmr1stablegeneralizabler1style}           & 3B  & 63.7 & 61.5 & 40.5 & 36.8 & 78.1 & 55.9 & 100  & 56.0 \\
    GroundingGPT\cite{li2024groundinggpt}     & 7B  & 49.6 & 41.4 & 16.4 & 16.1 & 37.3 & 52.3 & 100.0 & 47.0 \\
    Ferret\cite{you2023ferret}           & 7B  & 35.7 & 23.6 & 12.7 & 23.7 & 50.8 & 31.4 & 88.1  & 51.5 \\
    Ferret\cite{you2023ferret}           & 13B & 44.5 & 31.4 & 16.6 & 24.9 & 57.3 & 39.8 & 94.3  & 52.4 \\
    REESEEK\cite{jiang2025referring}         & 7B  & 46.2 & 40.7 & 35.1 & {--}   & {--}   & {--}   &   78.5  & {--} \\

    \bottomrule
  \end{tabular}
  \end{adjustbox}

  \caption{Comparison of baselines on KnowDR-Bench. In the positive sample, we summarize the model's performance in textual reasoning and visual grounding, and analyzed the impact of the success of textual reasoning on visual grounding. In the negative sample, we evaluated the performance under two settings.}
  \label{tab:1}
\end{table*}

\section{Results and Analysis}

\subsection{KnowDR-REC is Challenging for Current MLLMs}

Our benchmark results indicate that the KnowDR-REC task remains a significant challenge for current models. Specifically, all evaluated mainstream MLLMs fail to achieve satisfactory performance in the joint capability of textual reasoning and visual grounding. Some models (e.g., Gemini 2.5 Flash \cite{comanici2025gemini}, Grok-2 Vision \cite{xie2025maverix}, and Gemini 2.5 Pro \cite{comanici2025gemini}) demonstrate stronger language understanding but poor target localization performance. In contrast, smaller-scale models (e.g., Qwen2.5-VL-3B \cite{bai2025qwen25vltechnicalreport} and Qwen2.5-VL-7B \cite{bai2025qwen25vltechnicalreport}) exhibit stronger visual grounding abilities in zero-shot settings but weaker textual reasoning skills. Notably, models fine-tuned specifically for traditional REC tasks perform worse in our knowledge-intensive scenarios, suggesting limited generalization capability beyond the standard REC paradigm.


\subsection{Textual Reasoning and Visual Grounding are Tightly Coupled}

To better understand the interplay between textual reasoning and visual grounding, we introduced two conditional metrics: $A_{50}^{\text{match}}$ and $A_{50}^{\text{mismatch}}$. Across most models, we found a significant performance gap between these two metrics. Specifically, visual grounding accuracy was substantially higher when models successfully reasoned about entities in textual contexts. For example, the Qwen-VL-Max \cite{bai2023qwenvlversatilevisionlanguagemodel} achieved an accuracy of 80.4\% on the $A_{50}^{\text{match}}$ metric when the textual inference was correct, while the accuracy dropped sharply to 59.0\% on the $A_{50}^{\text{mismatch}}$ metric when textual reasoning failed. This clearly indicates that accurate textual reasoning enhances visual grounding capabilities in knowledge-driven REC scenarios.


\subsection{Models Lack Robustness Against Negative Samples}

We evaluate the capabilities of models in handling negative samples under two distinct settings and observe significant limitations in current mainstream MLLMs regarding their ability to reject or appropriately express uncertainty. In the Basic Setting, all evaluated models exhibit excessive confidence, frequently providing incorrect affirmations. Among these, only Qwen-VL-Max demonstrates relatively robust behavior, successfully rejecting approximately half of the negative samples. 

In the Instruction-guided Setting, negative samples are explicitly reformulated into a simplified binary classification task, where models are directly prompted to confirm or reject the existence of target entities. Surprisingly, most models perform near random-chance levels. However, a small subset of models, notably Gemini-2.5 Pro, Gemini 2.5 Flash, and Grok-2 Vision, exhibit robust performance with significantly higher binary classification accuracy. These results indicate: (1) MLLMs generally exhibit weak resistance to hallucinations when processing negative samples, and (2) the models' capability to recognize negative samples heavily depends on prompt-driven behavioral patterns induced by specific instructional contexts.

\begin{figure}[!t]
  \centering
  \includegraphics[width=0.43\textwidth]{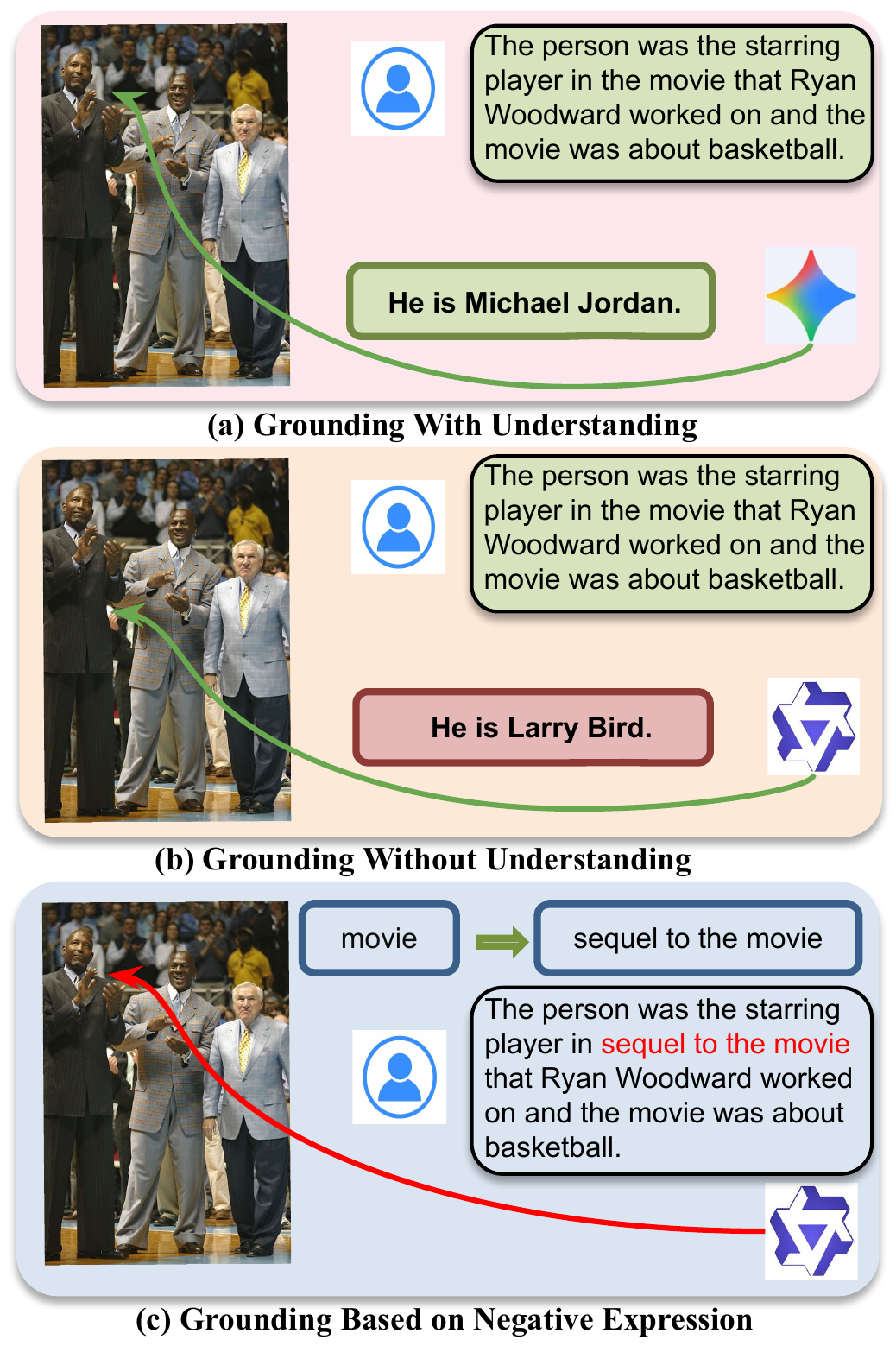}
  \caption{Visual grounding behaviors observed in REC tasks: (a) correct grounding based on correct understanding, (b) correct grounding based on incorrect understanding, and (c) “correct” grounding under negative descriptions.}
  \label{fig:4}
\end{figure}

\subsection{Shortcut Memorization Through Instance-level Visual Binding}

Although textual reasoning and visual grounding are tightly coupled, we observe that all models still demonstrate a certain degree of visual grounding capability even when they fail to perform textual reasoning. Moreover, many models perform significantly better in visual grounding than in textual reasoning, a phenomenon we refer to as \textit{Visual Grounding without Understanding}.

To further investigate this phenomenon, we analyze model outputs on negative samples under the Basic Setting, particularly focusing on cases where models still predict bounding boxes, as shown in Table~\ref{tab:2}. Surprisingly, results show that most models exhibit behavior on negative samples that is similar to their behavior on positive ones. Specifically, even when the input textual expression is incorrect and fails to refer to any instance in the image, models still exhibit a non-trivial ability to localize the target corresponding to the original positive sample, as shown in Figure~\ref{fig:4}(c). For instance, on Qwen-VL-Plus \cite{bai2023qwenvlversatilevisionlanguagemodel}, we find that the model achieves an Acc@50 of 64.6\% on correctly paired inputs, while still reaching 59.6\% on adversarial text-image pairs—a relatively small performance gap. Only Gemini 2.5 Pro, Gemini 2.5 Flash, and Grok-2 Vision identify the mismatch between textual expression and visual content.

Collectively, we speculate that memorization patterns play a critical role in visual grounding, and textual understanding is not the sole basis for instance selection. However, in the Referring Expression Comprehension (REC) task, language should serve as the only reliable cue for identification and localization. This fundamentally challenges the reliability and interpretability of current models.

\renewcommand{\arraystretch}{1.05}
\setlength{\tabcolsep}{3pt}

\begin{table}[htbp]
  \small
  \centering
  \begin{tabular}{lcccc}      
    \toprule
    \multirow{2}{*}{\textbf{Methods}} & \multirow{2}{*}{\textbf{Size}} & \multicolumn{3}{c}{\textbf{Negative Samples}} \\  
    \cmidrule(lr){3-5}                           
    & & \textbf{$\mathrm{Acc}_{0.5}$} & \textbf{$\mathrm{Acc}_{0.75}$} & \textbf{$\mathrm{Acc}_{0.9}$} \\
    \midrule
    \multicolumn{4}{l}{\textbf{Closed-Source Generalist MLLMs}}\\
    Gemini-2.5-Pro     & --  & 9.1 & 1.2 & 0.2 \\
    Gemini-2.5-Flash   & --  & 1.0 & 0.0 & 0.0 \\
    Grok-2-vision      & --  & 4.3 & 2.7 & 1.6 \\
    Qwen-VL-Plus       & --  & 59.6 & 53.1 & 24.8 \\
    Qwen-VL-Max       & --  & 56.9 & 53.1 & 36.7 \\
    \midrule
    \multicolumn{4}{l}{\textbf{Open-Source Generalist MLLMs}}\\
    Qwen2.5-VL      & 3B  & 52.1 & 48.8 & 34.2 \\
    Qwen2.5-VL      & 7B  & 25.3 & 18.3 & 9.1 \\
    Qwen2.5-VL     & 32B  & 46.5 & 36.3 & 21.3 \\
    Qwen2.5-VL    & 72B  & 61.9 & 55.9 & 41.3 \\
    SPHINX-Tiny       & 1.1B  & 6.2 & 1.1 & 0.0 \\
    SPHINX-v2-1k      & 13B  & 56.2 & 18.7 & 0.0 \\
    \midrule
    \multicolumn{4}{l}{\textbf{Open-Source Specialist MLLMs}}\\
    VLM-R1            & 3B   & 49.4 & 43.6 & 29.8 \\
    GroundingGPT      & 7B   & 49.6 & 38.4 & 14.9 \\
    Ferret            & 7B   & 32.3 & 19.1 & 7.9 \\
    Ferret             & 13B   & 46.6 & 29.7 & 11.5 \\
    REESEEK          & 7B   & 27.1 & 19.6 & 14.1 \\

    \bottomrule

  \end{tabular}
  \caption{Evaluation results on negative samples under the Basic Setting, computed by comparing predicted bounding boxes against the ground-truth boxes from corresponding positive samples.}
  \label{tab:2}
\end{table}

\begin{figure}[!t]
  \centering
  \includegraphics[width=0.5\textwidth]{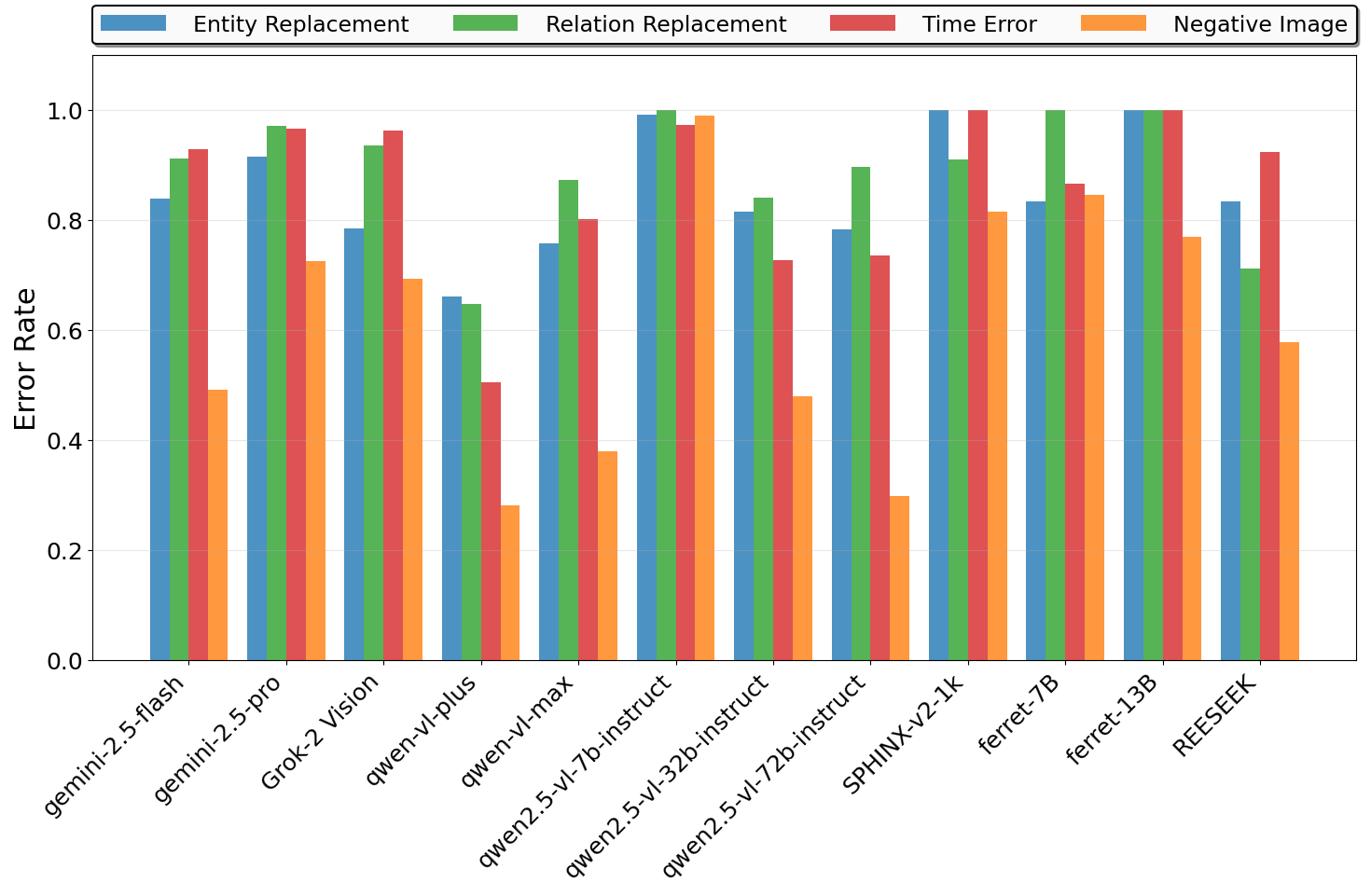}
  \caption{A comparison of error rates across multiple models under different types of negative samples.}
  \label{fig:5}
\end{figure}

\subsection{Model Sensitivity to Different Negative Sample Types}

To evaluate the impact of different types of negative samples on model performance, we conducted systematic analyses under the basic setting. As shown in Figure~\ref{fig:5}, most models exhibit relatively low error rates when handling negative images.

However, dealing with fine-grained negative samples derived from nuanced textual perturbations, such as entity replacement, relation replacement, and time error, all models show a significant drop in performance, revealing a notable limitation of current MLLMs: their limited ability to perform reasoning over nuanced textual expressions.

\section{Conclusion and Limitation}

In this study, we introduce KnowDR-REC, a novel benchmark designed to evaluate reasoning over external world knowledge in the referring expression comprehension (REC) task. The benchmark includes both positive and negative samples, with the negative samples created by perturbing entities, relations, or temporal information in the expression. Comprehensive experiments reveal key limitations of state-of-the-art multimodal large language models (MLLMs), including unreliable visual grounding evaluations, overreliance on memorization rather than genuine semantic understanding, inability to reject incorrect queries, and poor performance on knowledge-intensive REC tasks. These findings highlight the necessity of our proposed benchmark. Although the current version focuses on person entities due to the richness of available online information, it lays a structured foundation for advancing research toward more reliable, interpretable, and knowledge-intensive AI systems in complex real-world grounding scenarios.


\bibliography{aaai2026}

\begin{thebibliography}{43}
\providecommand{\natexlab}[1]{#1}

\bibitem[{Bai et~al.(2023)Bai, Bai, Yang, Wang, Tan, Wang, Lin, Zhou, and Zhou}]{bai2023qwenvlversatilevisionlanguagemodel}
Bai, J.; Bai, S.; Yang, S.; Wang, S.; Tan, S.; Wang, P.; Lin, J.; Zhou, C.; and Zhou, J. 2023.
\newblock Qwen-VL: A Versatile Vision-Language Model for Understanding, Localization, Text Reading, and Beyond.
\newblock arXiv:2308.12966.

\bibitem[{Bai et~al.(2025)Bai, Chen, Liu, Wang, Ge, Song, Dang, Wang, Wang, Tang, Zhong, Zhu, Yang, Li, Wan, Wang, Ding, Fu, Xu, Ye, Zhang, Xie, Cheng, Zhang, Yang, Xu, and Lin}]{bai2025qwen25vltechnicalreport}
Bai, S.; Chen, K.; Liu, X.; Wang, J.; Ge, W.; Song, S.; Dang, K.; Wang, P.; Wang, S.; Tang, J.; Zhong, H.; Zhu, Y.; Yang, M.; Li, Z.; Wan, J.; Wang, P.; Ding, W.; Fu, Z.; Xu, Y.; Ye, J.; Zhang, X.; Xie, T.; Cheng, Z.; Zhang, H.; Yang, Z.; Xu, H.; and Lin, J. 2025.
\newblock Qwen2.5-VL Technical Report.
\newblock arXiv:2502.13923.

\bibitem[{Bai et~al.(2024)Bai, Wang, Xiao, He, Han, Zhang, and Shou}]{bai2024hallucination}
Bai, Z.; Wang, P.; Xiao, T.; He, T.; Han, Z.; Zhang, Z.; and Shou, M.~Z. 2024.
\newblock Hallucination of multimodal large language models: A survey.
\newblock \emph{arXiv preprint arXiv:2404.18930}.

\bibitem[{Bridge(2001)}]{bridge2001wikipedia}
Bridge, A.-M. 2001.
\newblock Wikipedia, the free encyclopedia.
\newblock \emph{San Francisco (CA): Wikimedia Foundation}.

\bibitem[{Cai et~al.(2022)Cai, Xiang, Gao, Zhang, Li, and Li}]{cai2022temporal}
Cai, B.; Xiang, Y.; Gao, L.; Zhang, H.; Li, Y.; and Li, J. 2022.
\newblock Temporal knowledge graph completion: A survey.
\newblock \emph{arXiv preprint arXiv:2201.08236}.

\bibitem[{Cao et~al.(2020)Cao, Shi, Pan, Nie, Xiang, Hou, Li, He, and Zhang}]{cao2020kqa}
Cao, S.; Shi, J.; Pan, L.; Nie, L.; Xiang, Y.; Hou, L.; Li, J.; He, B.; and Zhang, H. 2020.
\newblock KQA pro: A dataset with explicit compositional programs for complex question answering over knowledge base.
\newblock \emph{arXiv preprint arXiv:2007.03875}.

\bibitem[{Chen et~al.(2024)Chen, Ma, Zhang, Xu, Qian, Yang, Fouhey, and Chai}]{chen2024multi}
Chen, X.; Ma, Z.; Zhang, X.; Xu, S.; Qian, S.; Yang, J.; Fouhey, D.; and Chai, J. 2024.
\newblock Multi-object hallucination in vision language models.
\newblock \emph{Advances in Neural Information Processing Systems}, 37: 44393--44418.

\bibitem[{Comanici et~al.(2025)Comanici, Bieber, Schaekermann, Pasupat, Sachdeva, Dhillon, Blistein, Ram, Zhang, Rosen et~al.}]{comanici2025gemini}
Comanici, G.; Bieber, E.; Schaekermann, M.; Pasupat, I.; Sachdeva, N.; Dhillon, I.; Blistein, M.; Ram, O.; Zhang, D.; Rosen, E.; et~al. 2025.
\newblock Gemini 2.5: Pushing the frontier with advanced reasoning, multimodality, long context, and next generation agentic capabilities.
\newblock \emph{arXiv preprint arXiv:2507.06261}.

\bibitem[{Hu et~al.(2023)Hu, Luan, Chen, Khandelwal, Joshi, Lee, Toutanova, and Chang}]{hu2023open}
Hu, H.; Luan, Y.; Chen, Y.; Khandelwal, U.; Joshi, M.; Lee, K.; Toutanova, K.; and Chang, M.-W. 2023.
\newblock Open-domain visual entity recognition: Towards recognizing millions of wikipedia entities.
\newblock In \emph{Proceedings of the IEEE/CVF International Conference on Computer Vision}, 12065--12075.

\bibitem[{Huang et~al.(2023)Huang, Dong, Wang, Hao, Singhal, Ma, Lv, Cui, Mohammed, Patra et~al.}]{huang2023language}
Huang, S.; Dong, L.; Wang, W.; Hao, Y.; Singhal, S.; Ma, S.; Lv, T.; Cui, L.; Mohammed, O.~K.; Patra, B.; et~al. 2023.
\newblock Language is not all you need: Aligning perception with language models.
\newblock \emph{Advances in Neural Information Processing Systems}, 36: 72096--72109.

\bibitem[{Huang, Zhang, and Satoh(2025)}]{huang2025resedis}
Huang, Z.; Zhang, Y.; and Satoh, S. 2025.
\newblock ReSeDis: A Dataset for Referring-based Object Search across Large-Scale Image Collections.
\newblock \emph{arXiv preprint arXiv:2506.15180}.

\bibitem[{Hurst et~al.(2024)Hurst, Lerer, Goucher, Perelman, Ramesh, Clark, Ostrow, Welihinda, Hayes, Radford et~al.}]{hurst2024gpt}
Hurst, A.; Lerer, A.; Goucher, A.~P.; Perelman, A.; Ramesh, A.; Clark, A.; Ostrow, A.; Welihinda, A.; Hayes, A.; Radford, A.; et~al. 2024.
\newblock Gpt-4o system card.
\newblock \emph{arXiv preprint arXiv:2410.21276}.

\bibitem[{Jiang et~al.(2025)Jiang, Wu, Zeng, Ren, Xiong, Chen, Liu, and Zhang}]{jiang2025referring}
Jiang, Q.; Wu, L.; Zeng, Z.; Ren, T.; Xiong, Y.; Chen, Y.; Liu, Q.; and Zhang, L. 2025.
\newblock Referring to any person.
\newblock \emph{arXiv preprint arXiv:2503.08507}.

\bibitem[{Li et~al.(2023)Li, Du, Zhou, Wang, Zhao, and Wen}]{li2023evaluating}
Li, Y.; Du, Y.; Zhou, K.; Wang, J.; Zhao, W.~X.; and Wen, J.-R. 2023.
\newblock Evaluating object hallucination in large vision-language models.
\newblock \emph{arXiv preprint arXiv:2305.10355}.

\bibitem[{Li et~al.(2024)Li, Xu, Zhang, Song, Cai, Qi, Zhou, Pan, Li, Vu et~al.}]{li2024groundinggpt}
Li, Z.; Xu, Q.; Zhang, D.; Song, H.; Cai, Y.; Qi, Q.; Zhou, R.; Pan, J.; Li, Z.; Vu, V.~T.; et~al. 2024.
\newblock Groundinggpt: Language enhanced multi-modal grounding model.
\newblock \emph{arXiv preprint arXiv:2401.06071}.

\bibitem[{Lin et~al.(2023)Lin, Liu, Zhang, Gao, Qiu, Xiao, Qiu, Lin, Shao, Chen et~al.}]{lin2023sphinx}
Lin, Z.; Liu, C.; Zhang, R.; Gao, P.; Qiu, L.; Xiao, H.; Qiu, H.; Lin, C.; Shao, W.; Chen, K.; et~al. 2023.
\newblock Sphinx: The joint mixing of weights, tasks, and visual embeddings for multi-modal large language models.
\newblock \emph{arXiv preprint arXiv:2311.07575}.

\bibitem[{Liu et~al.(2024{\natexlab{a}})Liu, Zhang, Qiu, Huang, Lin, Zhao, Geng, Lin, Jin, Zhang et~al.}]{liu2024sphinx}
Liu, D.; Zhang, R.; Qiu, L.; Huang, S.; Lin, W.; Zhao, S.; Geng, S.; Lin, Z.; Jin, P.; Zhang, K.; et~al. 2024{\natexlab{a}}.
\newblock Sphinx-x: Scaling data and parameters for a family of multi-modal large language models.
\newblock \emph{arXiv preprint arXiv:2402.05935}.

\bibitem[{Liu et~al.(2023)Liu, Li, Wu, and Lee}]{liu2023visual}
Liu, H.; Li, C.; Wu, Q.; and Lee, Y.~J. 2023.
\newblock Visual instruction tuning.
\newblock \emph{Advances in neural information processing systems}, 36: 34892--34916.

\bibitem[{Liu et~al.(2024{\natexlab{b}})Liu, Xue, Chen, Chen, Zhao, Wang, Hou, Li, and Peng}]{liu2024survey}
Liu, H.; Xue, W.; Chen, Y.; Chen, D.; Zhao, X.; Wang, K.; Hou, L.; Li, R.; and Peng, W. 2024{\natexlab{b}}.
\newblock A survey on hallucination in large vision-language models.
\newblock \emph{arXiv preprint arXiv:2402.00253}.

\bibitem[{Liu et~al.(2024{\natexlab{c}})Liu, Yang, Li, and Wang}]{liu2024finecops}
Liu, J.; Yang, X.; Li, W.; and Wang, P. 2024{\natexlab{c}}.
\newblock Finecops-ref: A new dataset and task for fine-grained compositional referring expression comprehension.
\newblock \emph{arXiv preprint arXiv:2409.14750}.

\bibitem[{Liu et~al.(2019)Liu, Liu, Bai, and Yuille}]{liu2019clevr}
Liu, R.; Liu, C.; Bai, Y.; and Yuille, A.~L. 2019.
\newblock Clevr-ref+: Diagnosing visual reasoning with referring expressions.
\newblock In \emph{Proceedings of the IEEE/CVF conference on computer vision and pattern recognition}, 4185--4194.

\bibitem[{Mao et~al.(2016)Mao, Huang, Toshev, Camburu, Yuille, and Murphy}]{mao2016generation}
Mao, J.; Huang, J.; Toshev, A.; Camburu, O.; Yuille, A.~L.; and Murphy, K. 2016.
\newblock Generation and comprehension of unambiguous object descriptions.
\newblock In \emph{Proceedings of the IEEE conference on computer vision and pattern recognition}, 11--20.

\bibitem[{Marino et~al.(2019)Marino, Rastegari, Farhadi, and Mottaghi}]{marino2019ok}
Marino, K.; Rastegari, M.; Farhadi, A.; and Mottaghi, R. 2019.
\newblock Ok-vqa: A visual question answering benchmark requiring external knowledge.
\newblock In \emph{Proceedings of the IEEE/cvf conference on computer vision and pattern recognition}, 3195--3204.

\bibitem[{Miyai et~al.(2024)Miyai, Yang, Zhang, Ming, Yu, Irie, Li, Li, Liu, and Aizawa}]{miyai2024unsolvable}
Miyai, A.; Yang, J.; Zhang, J.; Ming, Y.; Yu, Q.; Irie, G.; Li, Y.; Li, H.; Liu, Z.; and Aizawa, K. 2024.
\newblock Unsolvable problem detection: Robust understanding evaluation for large multimodal models.
\newblock \emph{arXiv preprint arXiv:2403.20331}.

\bibitem[{Puerto, {\c{S}}ahin, and Gurevych(2021)}]{puerto2021metaqa}
Puerto, H.; {\c{S}}ahin, G.~G.; and Gurevych, I. 2021.
\newblock Metaqa: Combining expert agents for multi-skill question answering.
\newblock \emph{arXiv preprint arXiv:2112.01922}.

\bibitem[{Qiao, Deng, and Wu(2020)}]{qiao2020referring}
Qiao, Y.; Deng, C.; and Wu, Q. 2020.
\newblock Referring expression comprehension: A survey of methods and datasets.
\newblock \emph{IEEE Transactions on Multimedia}, 23: 4426--4440.

\bibitem[{Schulter et~al.(2023)Schulter, Suh, Dafnis, Zhang, Zhao, Metaxas et~al.}]{schulter2023omnilabel}
Schulter, S.; Suh, Y.; Dafnis, K.~M.; Zhang, Z.; Zhao, S.; Metaxas, D.; et~al. 2023.
\newblock Omnilabel: A challenging benchmark for language-based object detection.
\newblock In \emph{Proceedings of the IEEE/CVF International Conference on Computer Vision}, 11953--11962.

\bibitem[{Shah et~al.(2019)Shah, Mishra, Yadati, and Talukdar}]{shah2019kvqa}
Shah, S.; Mishra, A.; Yadati, N.; and Talukdar, P.~P. 2019.
\newblock Kvqa: Knowledge-aware visual question answering.
\newblock In \emph{Proceedings of the AAAI conference on artificial intelligence}, volume~33, 8876--8884.

\bibitem[{Shen et~al.(2025)Shen, Liu, Li, Fang, Ma, Liao, Shen, Zhang, Zhao, Zhang, Xu, and Zhao}]{shen2025vlmr1stablegeneralizabler1style}
Shen, H.; Liu, P.; Li, J.; Fang, C.; Ma, Y.; Liao, J.; Shen, Q.; Zhang, Z.; Zhao, K.; Zhang, Q.; Xu, R.; and Zhao, T. 2025.
\newblock VLM-R1: A Stable and Generalizable R1-style Large Vision-Language Model.
\newblock arXiv:2504.07615.

\bibitem[{Sohan, Sai~Ram, and Rami~Reddy(2024)}]{sohan2024review}
Sohan, M.; Sai~Ram, T.; and Rami~Reddy, C.~V. 2024.
\newblock A review on yolov8 and its advancements.
\newblock In \emph{International Conference on Data Intelligence and Cognitive Informatics}, 529--545. Springer.

\bibitem[{Sun et~al.(2022)Sun, Fan, Guo, Zhang, and Cheng}]{sun2022visual}
Sun, W.; Fan, Y.; Guo, J.; Zhang, R.; and Cheng, X. 2022.
\newblock Visual named entity linking: A new dataset and a baseline.
\newblock \emph{arXiv preprint arXiv:2211.04872}.

\bibitem[{Talmor and Berant(2018)}]{talmor2018web}
Talmor, A.; and Berant, J. 2018.
\newblock The web as a knowledge-base for answering complex questions.
\newblock \emph{arXiv preprint arXiv:1803.06643}.

\bibitem[{Wang et~al.(2020)Wang, Liu, Li, and Wu}]{wang2020give}
Wang, P.; Liu, D.; Li, H.; and Wu, Q. 2020.
\newblock Give me something to eat: Referring expression comprehension with commonsense knowledge.
\newblock In \emph{Proceedings of the 28th ACM International Conference on Multimedia}, 28--36.

\bibitem[{Wang et~al.(2024)Wang, Lv, Yu, Hong, Qi, Wang, Ji, Yang, Zhao, XiXuan et~al.}]{wang2024cogvlm}
Wang, W.; Lv, Q.; Yu, W.; Hong, W.; Qi, J.; Wang, Y.; Ji, J.; Yang, Z.; Zhao, L.; XiXuan, S.; et~al. 2024.
\newblock Cogvlm: Visual expert for pretrained language models.
\newblock \emph{Advances in Neural Information Processing Systems}, 37: 121475--121499.

\bibitem[{Wei et~al.(2024)Wei, Zhao, Yan, Zhang, and Xu}]{wei2024large}
Wei, F.; Zhao, J.; Yan, K.; Zhang, H.; and Xu, C. 2024.
\newblock A large-scale human-centric benchmark for referring expression comprehension in the LMM era.
\newblock \emph{Advances in Neural Information Processing Systems}, 37: 69566--69587.

\bibitem[{Wei et~al.(2022)Wei, Wang, Schuurmans, Bosma, Xia, Chi, Le, Zhou et~al.}]{wei2022chain}
Wei, J.; Wang, X.; Schuurmans, D.; Bosma, M.; Xia, F.; Chi, E.; Le, Q.~V.; Zhou, D.; et~al. 2022.
\newblock Chain-of-thought prompting elicits reasoning in large language models.
\newblock \emph{Advances in neural information processing systems}, 35: 24824--24837.

\bibitem[{Wu et~al.(2020)Wu, Lin, Cohen, Bui, and Maji}]{wu2020phrasecut}
Wu, C.; Lin, Z.; Cohen, S.; Bui, T.; and Maji, S. 2020.
\newblock Phrasecut: Language-based image segmentation in the wild.
\newblock In \emph{Proceedings of the IEEE/CVF Conference on Computer Vision and Pattern Recognition}, 10216--10225.

\bibitem[{Xie et~al.(2025)Xie, Wei, Kuthiala, Zheng, Bal, Dabhi, Wen, Rustagi, Lai, Khyalia et~al.}]{xie2025maverix}
Xie, L.; Wei, G.~Z.; Kuthiala, A.; Zheng, C.; Bal, A.; Dabhi, M.; Wen, L.; Rustagi, T.; Lai, E.; Khyalia, S.; et~al. 2025.
\newblock MAVERIX: Multimodal Audio-Visual Evaluation Reasoning IndeX.
\newblock \emph{arXiv preprint arXiv:2503.21699}.

\bibitem[{Xu, Zhu, and Yang(2024)}]{xu2024mc}
Xu, Y.; Zhu, L.; and Yang, Y. 2024.
\newblock Mc-bench: A benchmark for multi-context visual grounding in the era of mllms.
\newblock \emph{arXiv preprint arXiv:2410.12332}.

\bibitem[{Yang et~al.(2018)Yang, Qi, Zhang, Bengio, Cohen, Salakhutdinov, and Manning}]{yang2018hotpotqa}
Yang, Z.; Qi, P.; Zhang, S.; Bengio, Y.; Cohen, W.~W.; Salakhutdinov, R.; and Manning, C.~D. 2018.
\newblock HotpotQA: A dataset for diverse, explainable multi-hop question answering.
\newblock \emph{arXiv preprint arXiv:1809.09600}.

\bibitem[{You et~al.(2023)You, Zhang, Gan, Du, Zhang, Wang, Cao, Chang, and Yang}]{you2023ferret}
You, H.; Zhang, H.; Gan, Z.; Du, X.; Zhang, B.; Wang, Z.; Cao, L.; Chang, S.-F.; and Yang, Y. 2023.
\newblock Ferret: Refer and ground anything anywhere at any granularity.
\newblock \emph{arXiv preprint arXiv:2310.07704}.

\bibitem[{Yu et~al.(2016)Yu, Poirson, Yang, Berg, and Berg}]{yu2016modeling}
Yu, L.; Poirson, P.; Yang, S.; Berg, A.~C.; and Berg, T.~L. 2016.
\newblock Modeling context in referring expressions.
\newblock In \emph{European conference on computer vision}, 69--85. Springer.

\bibitem[{Zang et~al.(2025)Zang, Li, Han, Zhou, and Loy}]{zang2025contextual}
Zang, Y.; Li, W.; Han, J.; Zhou, K.; and Loy, C.~C. 2025.
\newblock Contextual object detection with multimodal large language models.
\newblock \emph{International Journal of Computer Vision}, 133(2): 825--843.

\end{thebibliography}

\end{document}